\documentclass[letterpaper]{article} % DO NOT CHANGE THIS
\usepackage{aaai2026}  % DO NOT CHANGE THIS
\usepackage{times}  % DO NOT CHANGE THIS
\usepackage{helvet}  % DO NOT CHANGE THIS
\usepackage{courier}  % DO NOT CHANGE THIS
\usepackage[hyphens]{url}  % DO NOT CHANGE THIS
\usepackage{graphicx} % DO NOT CHANGE THIS
\usepackage{amsmath}
\usepackage{natbib}  % DO NOT CHANGE THIS AND DO NOT ADD ANY OPTIONS TO IT
\usepackage{caption} % DO NOT CHANGE THIS AND DO NOT ADD ANY OPTIONS TO IT
\usepackage{algorithm}
\usepackage{algorithmic}
\usepackage{comment}

\usepackage{newfloat}
\usepackage{listings}
\DeclareCaptionStyle{ruled}{labelfont=normalfont,labelsep=colon,strut=off} % DO NOT CHANGE THIS
\floatstyle{ruled}
\newfloat{listing}{tb}{lst}{}
\floatname{listing}{Listing}
\nocopyright

\title{
Bio-inspired Learning and Decision-Making\\ with Probabilistic In-Memory Computing Hardware: Part 1
}

\author{
Thomas Dalgaty\textsuperscript{\rm 1},
Eiji Kawasaki\textsuperscript{\rm 1},
Miguel de Prado\textsuperscript{\rm 2, \rm 3},
Devendra Vyas\textsuperscript{\rm 2},
Tommaso Salvatori\textsuperscript{\rm 2 \rm 4}
}

\affiliations{
\textsuperscript{\rm 1}CEA-List, Grenoble, France\\
\textsuperscript{\rm 2} formerly VERSES AI Research Lab \\
\textsuperscript{\rm 3} now PRAESC AI 
\textsuperscript{\rm 4} now TU Wien, Vienna, Austria
}

\begin{document}

\maketitle

\begin{abstract}
Learning and decision-making in animals are often modeled as Bayesian processes, where sensory evidence is integrated with prior beliefs to guide behavior in the face of uncertainty. But what are the inherent neural dynamics that give rise to this ability, and how could they be replicated in computing systems?
This abstract discusses a biologically grounded framework in which noisy neural and synaptic dynamics perform inference and learning via stochastic sampling from an internal energy function, capturing uncertainty over latent states and model parameters through neural and synaptic variability, respectively.
This enables approaches such as predictive coding networks to account for epistemic uncertainty via Markov chain Monte Carlo sampling.
Drawing a parallel between intrinsic noise in biological systems and electrical noise in emerging probabilistic analogue memory technologies, we highlight how analogue in-memory computing hardware naturally emerges as the solution for massively scalable and energy-efficient probabilistic inference.
\end{abstract}

% Uncomment the following to link to your code, datasets, an extended version or similar.
% You must keep this block between (not within) the abstract and the main body of the paper.
% \begin{links}
%     \link{Code}{https://aaai.org/example/code}
%     \link{Datasets}{https://aaai.org/example/datasets}
%     \link{Extended version}{https://aaai.org/example/extended-version}
% \end{links}

\section{Introduction}

Animals exhibit remarkable adaptability in uncertain and dynamic environments, often behaving in ways that approximate Bayesian inference~\cite{knill1996perception,knill2004bayesian,Paunov_2024}. This has led to the hypothesis that the brain performs probabilistic reasoning by integrating sensory evidence with prior expectations, guided by uncertainty. Energy-based modeling is a compelling framework for understanding this process, which views the brain as a physical system minimizing its internal free energy~\cite{hinton1986learning,friston2010free} to perform inference and learning efficiently.
This has inspired computational formulations, such as predictive coding, in which neural populations continuously generate predictions about sensory input and update them through the minimization of prediction errors, implicitly performing Bayesian inference~\cite{rao1999predictive,friston2005theory,salvatori2023brain}. Biological neural systems, such as the brain, are inherently noisy. Thermodynamic fluctuations, stochastic neurotransmitter release, and variability in synaptic transmission~\cite{fatt1950some,conti2004action,white2000channel} introduce randomness at both the neuronal and synaptic levels~\cite{buesing2011neural, kappel2015synaptic}. Rather than being detrimental however, this noise may serve a computational role by enabling sampling-based inference through neural and synaptic variability. Increasing evidence suggests that the brain leverages its intrinsic stochastic dynamics to perform Bayesian inference via mechanisms akin to Markov chain Monte Carlo (MCMC) sampling~\cite{hoyer2002interpreting,berkes2011spontaneous}.

In this abstract, we motivate a biologically inspired framework for learning and decision-making that uses noise as a computational resource. Specifically, we discuss the development of a new form of predictive coding \cite{oliviers2024learning, sennesh2024divide}, capable of modeling its epistemic uncertainty. While predictive coding has been extensively studied as a mechanism for perceptual inference, its treatment of epistemic uncertainty --- uncertainty about the parameters or structure of the generative model --- remains underexplored. In particular, most implementations i) treat synaptic weights in a deterministic fashion, neglecting the possibility that the weights themselves could represent probability distributions, or ii) propose a variational posterior over the weights \cite{tschantz2025bayesian}, not reflecting the persistent changes that result from biological noise according to synaptic sampling theory~\cite{kappel2015synaptic}.

We propose a pathway whereby applying stochastic updates to both neural and synaptic variables within an underlying energy function, inference and learning naturally emerge as biologically plausible Markov chain Monte Carlo sampling processes (\textit{Algorithmic Formulation}).
In addition, we draw a parallel between intrinsic biological noise in the brain and the electrical noise in semiconductor technologies (\textit{Bayesian Neuromorphic Hardware}). Just as MCMC sampling offers a biologically plausible mechanism for Bayesian inference in neural systems, it also provides a framework for leveraging the inherent variability of analogue neuromorphic hardware. Rather than being a limitation, this device-level noise can be harnessed as a computational asset, enabling scalable energy-efficient Bayesian learning and inference~\cite{dalgaty2021situ,lin2025deep}.

\section{Algorithm Formulation}
\label{sec:algo}

The unnormalized posterior probability distribution over the states of a population of neurons, $\theta$, with synaptic weights, $\omega$, given an observation $X$ can be expressed as:
\begin{equation}
\tilde{P}(\theta, \omega \mid X) = P(X \mid \theta, \omega) \cdot P(\theta, \omega).
\end{equation}
The Boltzmann distribution can then be used to relate this unnormalized posterior to the joint energy function of $\theta$, $\omega$ and $X$
\begin{equation}
\tilde{P}(\theta, \omega \mid X) = \exp\big(-E(\theta, \omega, X)\big).
\end{equation}
This, therefore, allows the log probability of the posterior to be written as the combination of two energy functions
\begin{equation}
\log \tilde{P}(\theta, \omega \mid X) = -E_{\text{L}}(X, \theta, \omega) - E_{\text{prior}}(\theta, \omega).
\end{equation}
Depending on the choice of energy function and prior, Markov chain Monte Carlo methods can be used to generate samples from the neuron state probability density functions, $p(\theta)$, and synaptic weight probability distributions, $p(\omega)$. In the case of a shallow, linear predictive coding network with weights $\omega$, latent variables (neurons) $\theta$, and prediction errors $e^{(t)} = X - \omega^{(t)}\theta^{(t)}$, Langevin dynamics can be used to draw these samples through the updates
\begin{align}
\Delta \theta^{(t)} &= \tau_\theta
\left[
\tfrac{1}{\sigma_X^2}\,\omega^{(t)\top} e^{(t)}
- \tfrac{1}{\sigma_\theta^2}(\theta^{(t)} - \mu_\theta)
\right]
+ \sqrt{2\tau_\theta}\,\epsilon_\theta^{(t)}, \nonumber \\
\Delta \omega^{(t)} &= \tau_\omega
\left[
\tfrac{1}{\sigma_X^2}\, e^{(t)}\theta^{(t)\top}
- \tfrac{1}{\sigma_\omega^2}\,\omega^{(t)}
\right]
+ \sqrt{2\tau_\omega}\,\epsilon_\omega^{(t)},
\label{eq:neurons_synapses}
\end{align}
where $\epsilon_\theta$ and $\epsilon_\omega$ denote Gaussian unit noise, $\tau_\theta$ and $\tau_\omega$ are the neural and synaptic time constants, $\mu_{\theta}$ is the prior mean of the latent neurons, and $\sigma_X^2$, $\sigma_\theta^2$ and $\sigma_\omega^2$ are the variances of the Gaussian likelihood, the latent prior and the zero-mean weight prior, respectively. Consistent with synaptic sampling theory, the weights evolve on a slower timescale than the neural activity, $\tau_\omega \ll \tau_\theta$ \cite{kappel2015synaptic}. When learning from a dataset rather than a single observation, the likelihood term of the weight update must be scaled to reflect the full dataset, as in stochastic gradient Langevin dynamics \cite{welling2011bayesian}, so that the weights sample the posterior given all observations.

A single well-mixed chain of these dynamics draws samples from the weight posterior and therefore captures epistemic uncertainty, but reading this uncertainty out requires waiting for the chain to mix over time. Running an ensemble of chains in parallel, each a particle walking over the same weight posterior, instead provides an instantaneous estimate of epistemic uncertainty through the variability across the ensemble. Local groups of neural circuits sharing the same structural motifs \cite{narayanan2005redundancy,yuste2024neuronal} could be a mechanism by which such ensembles are realized in biological neural networks.

Unlike similar works that rely on the Hopfield energy, we adopt the predictive coding energy. The popularity of the Hopfield energy is driven by two primary factors. First, its symmetric weights avoid the need to compute the transpose of the weight matrix, an operation whose biological implementation remains debated. Second, its dynamics have been extensively analyzed and are well understood. The predictive coding energy nevertheless offers several advantages: its dynamics are simpler, and in the shallow linear model considered here the steady state of the network can be obtained analytically. It also naturally operates in its generative formulation, implementing a hierarchical Gaussian model \cite{friston2005theory}. The transpose appearing in the neural update of equation~\ref{eq:neurons_synapses} does not pose a problem for analogue hardware, where crossbar arrays can be read in both the forward and reverse directions, although its biological plausibility remains an open question shared with predictive coding more broadly.

\vspace{-1ex}
\section{Probabilistic analogue in-memory computing}
\label{sec:hardware}

The inference and learning algorithm outlined in equation~\ref{eq:neurons_synapses} is appealing because of the resonance with neural and synaptic sampling theories.
However, it is based on running an ensemble of parallel Langevin chains over the same posterior. The execution of MCMC is notoriously slow on conventional computer architectures, making the algorithmic latency required to perform this over a large ensemble quickly prohibitive. For example, in deterministic predictive coding models, a solution on a small network can be reached in about $15$ iterations \cite{salvatori2022learning}, while a similar model with Langevin dynamics needs more than $200$ \cite{oliviers2024learning}.

Just as biological systems are rich with intrinsic noise, so too are the semiconducting technologies used to build computing hardware. Thermal noise \cite{johnson1928thermal,nyquist1928thermal} arises from the Brownian agitation of conducting particles, shot noise \cite{schottky1918spontane} from quanta of particles overcoming energy barriers, and analogue memory devices exhibit programming variability due to the stochastic nature of mechanisms like conductive filament formation \cite{dalgaty2021situ,lin2025deep} or magnetic spin change \cite{dalgaty2023scaling}.
These physical phenomena have already been mapped onto efficient implementations of MCMC sampling, such as Metropolis-adjusted samplers and Langevin dynamics. They are particularly efficient because analogue memory devices can be used to draw samples from probability distributions, store these samples, and then perform vector arithmetic using these stored samples, all within the memory circuit. This is in contrast to GPUs, where parameters must be loaded in from an external high-bandwidth DRAM memory, subject to a limited bandwidth. This bandwidth imposes a sequential execution on an algorithm that is intrinsically massively-parallelizable. In-memory computing is, by contrast, structurally a massively parallel computing approach and offers the potential of speeding up the execution of energy-based models by several orders of magnitude.

\section{Research Outlook}
Within the European network of excellence dAIEDGE, we aim to advance the probabilistic analogue in-memory computing paradigm by showing how it can enable a massive scaling-up of energy-based models such as predictive coding. Such hardware can reduce the computational cost of explicitly modeling epistemic uncertainty, opening up new possibilities for frameworks such as active inference, where actions are guided by uncertainty-driven exploration.
Our immediate focus is twofold. First, we will investigate the performance of our proposed Bayesian predictive coding network for learning and making decisions within the energy-based framework. Second, we will explore how the algorithm detailed in this abstract can be efficiently mapped onto the intrinsic stochastic physics of analogue in-memory computing hardware, which naturally reflects the variability observed in biological systems.

\section{Acknowledgments}

This collaboration was supported by Horizon Europe's dAIEDGE network of excellence - Grant Agreement Number 101120726.

\bibliography{aaai2026,bib2}

@article{dalgaty2021situ,
  title={In situ learning using intrinsic memristor variability via Markov chain Monte Carlo sampling},
  author={Dalgaty, Thomas and Castellani, Niccolo and Turck, Cl{\'e}ment and Harabi, Kamel-Eddine and Querlioz, Damien and Vianello, Elisa},
  journal={Nature Electronics},
  volume={4},
  number={2},
  pages={151--161},
  year={2021},
  publisher={Nature Publishing Group UK London}
}

@article{nyquist1928thermal,
  title={Thermal agitation of electric charge in conductors},
  author={Nyquist, Harry},
  journal={Physical review},
  volume={32},
  number={1},
  pages={110},
  year={1928},
  publisher={APS}
}

@article{lin2025deep,
  title={Deep Bayesian active learning using in-memory computing hardware},
  author={Lin, Yudeng and Gao, Bin and Tang, Jianshi and Zhang, Qingtian and Qian, He and Wu, Huaqiang},
  journal={Nature Computational Science},
  volume={5},
  number={1},
  pages={27--36},
  year={2025},
  publisher={Nature Publishing Group US New York}
}

@article{salvatori2022learning,
  title={Learning on arbitrary graph topologies via predictive coding},
  author={Salvatori, Tommaso and Pinchetti, Luca and Millidge, Beren and Song, Yuhang and Bao, Tianyi and Bogacz, Rafal and Lukasiewicz, Thomas},
  journal={Advances in neural information processing systems},
  volume={35},
  pages={38232--38244},
  year={2022}
}

@article{kappel2015synaptic,
  title={Synaptic sampling: A Bayesian approach to neural network plasticity and rewiring},
  author={Kappel, David and Habenschuss, Stefan and Legenstein, Robert and Maass, Wolfgang},
  journal={Advances in neural information processing systems},
  volume={28},
  year={2015}
}

@article{friston2010free,
  title={The free-energy principle: a unified brain theory?},
  author={Friston, Karl},
  journal={Nature reviews neuroscience},
  volume={11},
  number={2},
  pages={127--138},
  year={2010},
  publisher={Nature publishing group}
}

@article{hinton1986learning,
  title={Learning and relearning in Boltzmann machines},
  author={Hinton, Geoffrey E and Sejnowski, Terrence J and others},
  journal={Parallel distributed processing: Explorations in the microstructure of cognition},
  volume={1},
  number={282-317},
  pages={2},
  year={1986}
}

@article{tschantz2025bayesian,
  title={Bayesian Predictive Coding},
  author={Tschantz, Alexander and Koudahl, Magnus and Linander, Hampus and Da Costa, Lancelot and Heins, Conor and Beck, Jeff and Buckley, Christopher},
  journal={arXiv preprint arXiv:2503.24016},
  year={2025}
}

@article{Paunov_2024, title={Multiple and subject-specific roles of uncertainty in reward-guided decision-making}, url={http://dx.doi.org/10.7554/eLife.103363.1}, DOI={10.7554/elife.103363.1}, publisher={eLife Sciences Publications, Ltd}, author={Paunov, Alexander and L’Hôtellier, Maëva and Guo, Dalin and He, Zoe and Yu, Angela and Meyniel, Florent}, year={2024}, month=dec }

@article{narayanan2005redundancy,
  title={Redundancy and synergy of neuronal ensembles in motor cortex},
  author={Narayanan, Nandakumar S and Kimchi, Eyal Y and Laubach, Mark},
  journal={Journal of Neuroscience},
  volume={25},
  number={17},
  pages={4207--4216},
  year={2005},
  publisher={Society for Neuroscience}
}

@article{yuste2024neuronal,
  title={Neuronal ensembles: Building blocks of neural circuits},
  author={Yuste, Rafael and Cossart, Rosa and Yaksi, Emre},
  journal={Neuron},
  volume={112},
  number={6},
  pages={875--892},
  year={2024},
  publisher={Elsevier}
}

@inproceedings{welling2011bayesian,
  title={Bayesian learning via stochastic gradient Langevin dynamics},
  author={Welling, Max and Teh, Yee W},
  booktitle={Proceedings of the 28th international conference on machine learning (ICML-11)},
  pages={681--688},
  year={2011}
}

@article{rao1999predictive,
  title={Predictive coding in the visual cortex: {A} functional interpretation of some extra-classical receptive-field effects},
  author={Rao, Rajesh P. N. and Ballard, Dana H.},
  journal={Nature Neuroscience},
  volume={2},
  number={1},
  pages={79--87},
  year={1999},
  publisher={Nature Publishing Group}
}

@article{salvatori2023brain,
  title={Brain-inspired computational intelligence via predictive coding},
  author={Salvatori, Tommaso and Mali, Ankur and Buckley, Christopher L and Lukasiewicz, Thomas and Rao, Rajesh PN and Friston, Karl and Ororbia, Alexander},
  journal={arXiv preprint arXiv:2308.07870},
  volume={13},
  year={2023},
  publisher={no}
}

@article{sennesh2024divide,
  title={Divide-and-Conquer Predictive Coding: A structured {B}ayesian inference algorithm},
  author={Sennesh, Eli and Wu, Hao and Salvatori, Tommaso},
  journal={arXiv preprint arXiv:2408.05834},
  year={2024}
}

@article{oliviers2024learning,
  title={Learning probability distributions of sensory inputs with {Monte Carlo} predictive coding},
  author={Oliviers, Gaspard and Bogacz, Rafal and Meulemans, Alexander},
  journal={PLoS Computational Biology},
  volume={20},
  number={10},
  pages={e1012532},
  year={2024},
  publisher={Public Library of Science San Francisco, CA USA}
}

@article{friston2005theory,
  title={A theory of cortical responses},
  author={Friston, Karl},
  journal={Philosophical Transactions of the Royal Society B: Biological Sciences},
  volume={360},
  number={1456},
  year={2005},
  publisher={The Royal Society London}
}

@inproceedings{dalgaty2023scaling,
  title={Scaling-up Memristor Monte Carlo with magnetic domain-wall physics},
  author={Dalgaty, Thomas and Yamada, Shogo and Molnos, Anca and Kawasaki, Eiji and Mesquida, Thomas and Rummens, Fran{\c{c}}ois and Shibata, Tatsuo and Urakawa, Yukihiro and Terasaki, Yukio and Sasaki, Tomoyuki and others},
  booktitle={MLNCP2023-37th NeurIPS Machine Learning with New Compute Paradigms workshop},
  year={2023}
}

@article{white2000channel,
  title={Channel noise in neurons},
  author={White, John A and Rubinstein, Jay T and Kay, Alan R},
  journal={Trends in neurosciences},
  volume={23},
  number={3},
  pages={131--137},
  year={2000},
  publisher={Elsevier}
}

@article{fatt1950some,
  title={Some observations on biological noise},
  author={Fatt, P and Katz, B},
  journal={Nature},
  volume={166},
  number={4223},
  pages={597--598},
  year={1950},
  publisher={Nature Publishing Group UK London}
}

@article{conti2004action,
  title={Action potential-evoked and ryanodine-sensitive spontaneous Ca2+ transients at the presynaptic terminal of a developing CNS inhibitory synapse},
  author={Conti, Rossella and Tan, Yusuf P and Llano, Isabel},
  journal={Journal of Neuroscience},
  volume={24},
  number={31},
  pages={6946--6957},
  year={2004},
  publisher={Soc Neuroscience}
}

@article{buesing2011neural,
  title={Neural dynamics as sampling: a model for stochastic computation in recurrent networks of spiking neurons},
  author={Buesing, Lars and Bill, Johannes and Nessler, Bernhard and Maass, Wolfgang},
  journal={PLoS computational biology},
  volume={7},
  number={11},
  pages={e1002211},
  year={2011},
  publisher={Public Library of Science San Francisco, USA}
}

@article{berkes2011spontaneous,
  title={Spontaneous cortical activity reveals hallmarks of an optimal internal model of the environment},
  author={Berkes, Pietro and Orb{\'a}n, Gerg{\H{o}} and Lengyel, M{\'a}t{\'e} and Fiser, J{\'o}zsef},
  journal={Science},
  volume={331},
  number={6013},
  pages={83--87},
  year={2011},
  publisher={American Association for the Advancement of Science}
}

@article{knill2004bayesian,
  title={The Bayesian brain: the role of uncertainty in neural coding and computation},
  author={Knill, David C and Pouget, Alexandre},
  journal={TRENDS in Neurosciences},
  volume={27},
  number={12},
  pages={712--719},
  year={2004},
  publisher={Elsevier}
}

@article{hoyer2002interpreting,
  title={Interpreting neural response variability as Monte Carlo sampling of the posterior},
  author={Hoyer, Patrik and Hyv{\"a}rinen, Aapo},
  journal={Advances in neural information processing systems},
  volume={15},
  year={2002}
}

@book{knill1996perception,
  title={Perception as Bayesian inference},
  author={Knill, David C and Richards, Whitman},
  year={1996},
  publisher={Cambridge University Press}
}

@article{schottky1918spontane,
  title={{\"U}ber spontane Stromschwankungen in verschiedenen Elektrizit{\"a}tsleitern},
  author={Schottky, Walter},
  journal={Annalen der physik},
  volume={362},
  number={23},
  pages={541--567},
  year={1918},
  publisher={WILEY-VCH Verlag Leipzig}
}

@article{johnson1928thermal,
  title={Thermal agitation of electricity in conductors},
  author={Johnson, John Bertrand},
  journal={Physical review},
  volume={32},
  number={1},
  pages={97},
  year={1928},
  publisher={APS}
}

\end{document}